\pdfoutput=1

\documentclass[11pt]{article}

\usepackage[final]{acl}

\usepackage{times}
\usepackage{latexsym}
\usepackage[section]{placeins}
\usepackage[T1]{fontenc}

\usepackage[utf8]{inputenc}

\usepackage{microtype}

\usepackage{mathtools}
\usepackage{booktabs} 
\usepackage{multirow}
\usepackage{scalerel} 
\usepackage{graphicx}
\graphicspath{    
    {./figs/} 
}

\usepackage{algorithm} 
\usepackage{algorithmic}  
\usepackage[algo2e]{algorithm2e}

\usepackage{tabularx}
\usepackage{soul}
\definecolor{lightblue}{rgb}{.88,.98,1}
\definecolor{lavendermist}{rgb}{0.9, 0.9, 0.98}
\definecolor{lemonchiffon}{rgb}{1.0, 0.98, 0.8}
\sethlcolor{lemonchiffon} 
\usepackage{wrapfig,lipsum}
\newcommand\mC[1]{\multicolumn{1}{c}}

\usepackage{titlesec}
\titleformat{\section}{\normalfont\large\bfseries\center}{\thesection.}{1em}{}
\titleformat{\subsection}{\normalfont\fontsize{11}{12}\bfseries\raggedright}{\thesubsection.}{1em}{}
\titleformat{\subsubsection}{\normalfont\normalsize\bfseries\raggedright}{\thesubsubsection.}{1em}{}
\renewcommand\thesection{\arabic{section}}
\renewcommand\thesubsection{\thesection.\arabic{subsection}}
\renewcommand\thesubsubsection{\thesubsection.\arabic{subsubsection}}

\titleformat*{\paragraph}{\bf\itshape}

\usepackage{epstopdf}
\usepackage[utf8]{inputenc}

\usepackage{hyperref}
\usepackage{xstring}

\usepackage{array}
\newcolumntype{P}[1]{>{\centering\arraybackslash}p{#1}}

\usepackage{color}
\usepackage{csquotes}

\usepackage{amsmath}
\usepackage{pifont}

\title{Perturbation-based Active Learning for Question Answering}

\author{Fan Luo \\
  University of Arizona \\
  Tucson, AZ, USA \\ 
  \texttt{fanluo@email.arizona.edu} \\\And
  Mihai Surdeanu \\
  University of Arizona \\
  Tucson, AZ, USA \\ 
  \texttt{msurdeanu@email.arizona.edu} \\}

\begin{document}
\maketitle
\begin{abstract}

Building a question answering (QA) model with less annotation costs can be achieved by utilizing active learning (AL) training strategy. It selects the most informative unlabeled training data to update the model effectively. Acquisition functions for AL are used to determine how informative each training example is, such as uncertainty or diversity based sampling. 
In this work, we propose a perturbation-based active learning acquisition strategy and demonstrate it is more effective than existing commonly used strategies.

\end{abstract}

\section{Introduction}

Contemporary state-of-the-art QA models are deep neural networks. Although employing deep models has led to impressive performance improvement, but they are data-hungry. Training and even fine-tuning such models requires a large amount of high-quality annotated training data to achieve a reasonably good performance. For question answering tasks, unlabeled data are easily obtained, but the answer labels are time-consuming and expensive to obtain. With a limited annotation budget in practice, selecting unlabeled training instances that are most informative to annotate is an effective way to reduce the annotation effort. The paradigm of choosing the most informative data to label is referred to as active learning.  

Active learning (AL) is a learning method that aims at minimizing labeling costs of training data acquisition without sacrificing accuracy \citep{fu2013survey}.  
With a limited annotation budget in practice, AL selects the most informative instances and queries their labels through the interaction with oracles (annotated by experts or apply crowd-sourcing techniques) to learn. In AL, the learner iteratively selects the most informative data that are helpful in evolving the model for annotation and updating the model with respect to the new annotations. Common acquisition functions for active learning are based on uncertainty and diversity. \citep{settles2009active,dor2020active,schroder2022revisiting}. The uncertainty-based function select difficult examples according to probability scores from the model’s predictions, and the diversity-based function selects heterogeneous data points in the feature space. The two approaches are orthogonal to each other, since uncertainty sampling is usually based on the model’s output, while diversity exploits information from the input feature space.\citep{dasgupta2011two} In this work, we propose PAL (Perturbation Active Learning) acquisition method, utilizing both the input feature and model’s output predictions to select the most informative instances. We conduct experiments to compare PAL with multiple AL strategies by fine-tuning pre-trained language model BERT for answer prediction and evaluate on the SQuAD dataset\citep{rajpurkar2016squad}.
 
\section{Related Works}
\subsection{Answer Extraction with Reader Model}
Widely adopted QA approaches use a two-stage retriever-reader pipeline \citep{chen2017reading}: a retriever first gathers relevant passages as evidence candidates, then a reader performs reading comprehension (RC) to extract an answer from the retrieved candidates to form an answer.  
\begin{figure*}
\centering
\includegraphics[width=16cm]{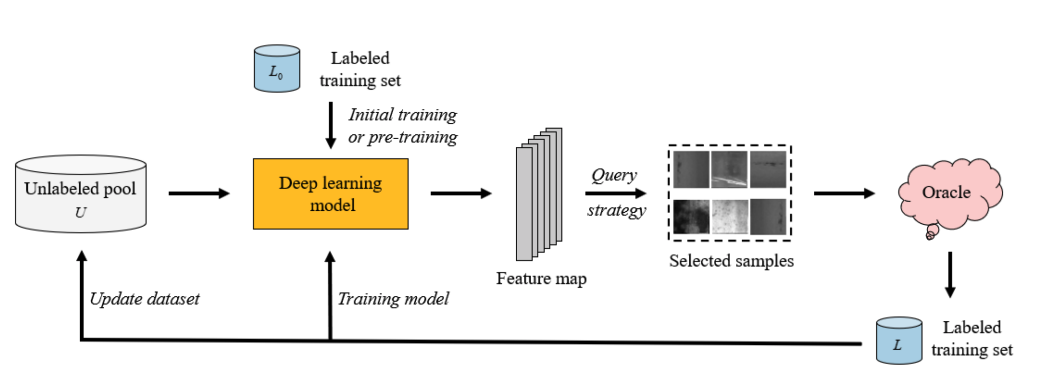}
\caption{A typical example of DeepAL model architecture (\citep{ren2021survey})}
\label{fig:DeepAL}
\end{figure*}
Typical Reading Comprehension (RC) task is a supervised learning problem that learns a predictor which takes a question and an input passage of text as inputs and gives the answer as output. Broadly, RC systems are grouped into three categories, namely rule-based, machine learning based and deep learning based models. Rule-based models are based
on hand-crafted rules that require substantial human effort, and are
incapable of generalization. A rule can be a logical combination of any of the language processing tasks such as part-of-speech tagging, semantic class tagging, and entity recognition. Traditional ML models 
transform RC into a supervised learning task, relying on a set of pre-defined features.
After designing a set of features to capture the information that helps to distinguish answer sentences from non-answer sentences, the learning algorithm generate a classifier
from the training examples for answer sentence classification. 
While these methods do not need to design 
hand-crafted rules, 
feature engineering 
is a critical necessity,
demanding 
scientists to provide task-driven and insightful feature vector representations for underlying optimization problems.
Besides, most traditional machine learning methods are also `data hungry', and could be potentially benefit from AL.
Lately, the research progress of RC has diverted from pure machine learning based models to deep learning based models \citep{chen2017reading,vakulenko2019message,reddy2019coqa,lei2018tvqa}. 
DL-based RC, also called neural reading comprehension, has made significant progress in recent years.
Neural RC models automatically generate representations to better extract contextual information and dramatically
with attention mechanism \citep{weston2014memory,antol2015vqa,xiong2016dynamic}.
The neural attention mechanism allows the system to focus on the most relevant part of the context paragraph in order to answer a question.
The attention mechanisms help improve the ability of RC systems to model complex interactions between a context paragraph and a query.  
A real game-changer is a new neural network architecture, called Transformer, introduced in the paper {\em ``Attention is All You Need"} \citep{vaswani2017attention}.
It is a type of neural network architecture that essentially utilizes the concept of self-attention. The self-attention layer allows each word position in an input sequence to attend to all positions in the sequence, which captures long-range dependencies between words, such as syntactic, semantic, and coreference relations. The transformer also performs multi-head attention, which allows the model to jointly capture different attentions from different sub-spaces, e.g., jointly attend to information that might indicate both coreference and syntactic relations \citep{storks2019recent}.

\subsection{Learning with Limited Annotations}

Effective training of deep QA models with very little labeled data can be achieved by either semi-supervised learning (SSL) or active learning (AL).

\subsubsection{Active Learning}
AL is a sub-field of machine learning (ML) in which a learning algorithm aims to achieve good accuracy with fewer training samples by interactively querying the oracles to label new data points \citep{zhan2021comparative}.  
AL aims to concentrate the expensive labeling process on the most informative instances 
from a (typically large) pool of unlabeled data 
in order to reach high accuracy with low-cost data labeling. 
AL has been shown to be effective in reducing the amount of labeling effort involved in training machine learning models. A good summary of active
learning works prior to the advances in DL can be found in \cite{settles2009active}.
With the impressive breakthroughs DL made in various challenging tasks,
the combination of DL and AL, referred as deep active learning (DeepAL), attracted widespread research interest in recent years.
Both DL and AL are sub-fields of machine learning. 
DL is also called representation learning. It realizes the automatic extraction of data features. DL has strong learning capabilities due to its complex structure, but this also means that DL requires a large number of labeled samples to complete the corresponding training. DL is limited by the high cost of sample labeling in some professional fields that require rich knowledge. 
AL focuses on the study of datasets, and it is also known as query learning. An effective AL algorithm can theoretically achieve exponential acceleration in labeling efficiency. 
Therefore, the combination of DL and AL is expected to
consider the complementary advantages of the two methods to
achieve superior results \citep{ren2021survey}. Figure~\ref{fig:DeepAL} illustrates a typical example of DeepAL model architecture.

\subsubsection{Semi-supervised learning}

SSL is a type of machine learning that uses both labeled and unlabeled data to create a model.
Whereas the collection of data is often cheap, labeling data can usually only be achieved at enormous costs because experts have to annotate the data manually. SSL combines supervised learning and the usage of unlabeled data \citep{reinders2019learning}. 
It is useful when there is not enough labeled data to create a model, but there is enough unlabeled large unlabeled data to provide some information about the model, which is a common case in many real-world machine learning problems.
SSL methods offer a wide range of methods for leveraging unlabeled data when learning prediction models. 
Most
of the SSL methods are based
on combinations of the supervised loss  
and an unsupervised loss. 
Classical SSL algorithms include the EM based algorithms, self-training, co-training, semi-supervised SVM (S3VM), graph-based methods, and boosting based SSL methods \citep{armoogum2019big}. A batch of novel models have
been recently introduced for SSL based on representation learning techniques,
such as generative models,
ladder networks and graph
embeddings.

\section{Methodology}

In this work, we introduce an AL acquisition strategy, Perturbation Active Learning (PAL), considering the robustness of the model as a signal of informativeness of active learning.
For the investigation of the effectiveness of PAL in combination with BERT \citep{devlin2018bert} for QA task, we conduct an empirical study to compare it with several commonly used AL query strategies with experiments by fine-tuning BERT-based model and evaluate on the SQuAD dataset \citep{rajpurkar2016squad}.

\subsection{Fine-tuning Reader Model}
 
A reader model performing reading comprehension (RC) to extract an answer from the earlier retrieved potentially relevant documents. It requires the model have a semantic understanding of the question and context, and be able to locate the position of an answer span. 
Training a transformer-based model from scratch is computationally expensive.
A pre-trained transformer model that has been already trained on massive amounts of text data (for example, BERT was trained on both BooksCorpus and Wikipedia) allows it to understand the language well and capture a surprising amount of common knowledge, which is crucial for the QA task. And with small changes in their architecture and pre-training objective, they can be fine-tuned to perform various types of downstream NLP tasks. 
Pre-trained transformer models can be treated as representations encoders by simply dropping the original output decision layer, with the learned embedding representations of the input text encoded in the final hidden state vectors.
Fine-tuning a pre-trained model on question answering dataset takes the concatenation of question and context sentences as a whole input sequence, and utilize the cross-attention between them to predict the correct answer span. Two probability scores are calculated for each context token, as the likelihood of the token being the start and end of the answer span. To compute the probability of each possible answer span, it is common to assume that the probability of a span being the answer can be estimated by the probability of tokens of being the start and end of the answer. The highest scored span is considered an answer. The span with the highest probability value (i.e., sum of start token and end token probabilities) is predicted as the answer. 

To build a high-performance QA model with less amount of annotated examples, we explore different acquisition strategies used by active learning to intelligently select question answer examples to label and learn during the next iteration. 

\subsection{Active Learning Iteration}

An active learner iteratively selects examples for annotation between rounds of training. As shown in Algorithm \ref{algorithm:AL-loop}, initially, one percent of the dataset is randomly selected to be annotated as $\mathcal{D}_{(0)}l$ for fine-tuning the already pre-trained BERT-BASE model, and all the rest is a large unlabeled data pool $\mathcal{D}_{(0)}u$.
Then an active learner iteratively selects examples for annotation between rounds of training.  
Assume in the current AL iteration $t$, labeled data with answer annotation for training is $\mathcal{D}_{(t)}l$ and a pool of unlabeled questions is $\mathcal{D}_{(t)}u$.
After continuing the fine-tune of the pre-trained model on $\mathcal{D}_{(t)}l$, the active learner applies the acquisition function to acquire a subset $\mathcal{D}_{(t)}al$, consisting of 10\% of unlabeled questions from the rest of the unlabeled dataset$\mathcal{D}_{(t)}u$. The newly acquired questions are then labeled, and so they are removed from $\mathcal{D}_u$ and are added to the labeled dataset $\mathcal{D}_l$ along with their gold answer: $\mathcal{D}_{(t+1)})l$ = $\mathcal{D}_{(t)}l \cup \mathcal{D}_{(t)}al$, and $\mathcal{D}_{(t+1)}u = \mathcal{D}_{(t)}u - \mathcal{D}_{(t)}al$. This procedure continues until all unlabeled questions are exhausted.

\begin{algorithm}
\SetArgSty{textup}
\DontPrintSemicolon
\KwIn{Unlabeled data pool $\mathcal{U}$, AL acquisition function $\phi(\cdot,\cdot,\cdot)$, AL selected unlabeled samples $\mathcal{D}_{(t)}al$}
\While{$\lvert\mathcal{D}_{(t)}u\rvert > 0$}{
  $\mathcal{M}_{(t)} \gets$ ~ Continue fine-tuning $M_{t-1}$ with $\mathcal{D}_{(t)}al$\\
  $\mathcal{D}_{(t)}al \longleftarrow \text{arg} \max_{x \in \mathcal{D}_{(t)}u} \phi(M_t, x, 10\%)$\\
  $\mathcal{D}_{(t+1)}l \longleftarrow \mathcal{D}_{(t)}l \cup label(\mathcal{D}_{(t)}al)$\\
  $\mathcal{D}_{(t+1)}u \longleftarrow \mathcal{D}_{(t)}u \setminus \mathcal{D}_{(t)}al$
}
\caption{Active Learning Approach }
\label{algorithm:AL-loop}
\end{algorithm}

\subsection{Acquisition Strategies}
Instead of randomly choosing examples from unlabelled corpus, AL acquisition strategies formulate an acquisition function that measures the usefulness of unlabeled data based on specified criteria. According to the results of the acquisition function, most informative unlabeled examples are then selected for annotation.

\subsubsection{Common Active Learning strategies}
\begin{figure*}
\centering
\includegraphics[width=14cm]{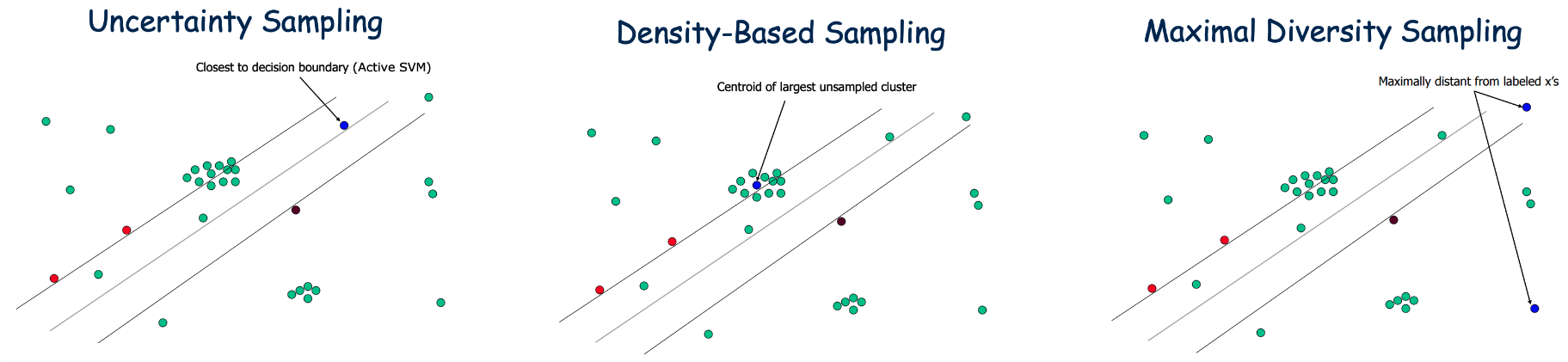}
\vspace{0.1in}
\caption{Common Active Learning Strategies (\citep{balcan2015active}) }
\label{fig:AL_Strategies}
\end{figure*}

The comparison of several common AL acquisition strategies used to select informative unlabeled data samples during the active learning process is illustrated in Figure~\ref{fig:AL_Strategies}.  

\paragraph{Least Confidence} This strategy selects instances with least confidence based on the output predictions of the model. The confidence score is assigned by the acquisition function to each unlabelled question as a measurement of model's uncertainty. For example, in a binary classification task, the instances with probability value around 0.5 are the ones with least confidence and the model highly uncertain about. For the QA task, the model outputs the candidate span with the highest probability value as the predicted answer for each question. The probability value of the predicted answer is considered as the confidence score for the question. The least confidence strategy selects questions whose predicted answer span has the least probability value to annotate.

\paragraph{Clustering} The clustering acquisition strategy feeds the BERT output embedding of the fine-tuned model for all the unlabeled questions concatenated with their context in Du into clustering algorithm such as K-Means. The k selected questions for annotation are sampled proportional from each cluster per the size of the cluster.

\paragraph{Maximal Diversity} It projects both unlabelled questions and labelled questions to the feature space using the embedding of the fine-tuned model. 
The goal is to query the unlabelled questions that have maximal distant from the labeled ones. Considering the heterogeneous of the labeled questions, we cluster the labeled questions in $\mathcal{D}_{(lab)}$, and return unlabelled questions with maximum distance to its closest labelled question centroid.

\subsubsection{Our Strategy: Perturbation-based AL} 

We propose this acquisition strategy in considering the robustness of the model as a signal of informativeness. We hypothesize that a robust model should product similar probability distributions on the part of the original context after perturbation the context with an additional distracting sentence. From another aspect, among unlabelled questions Du, the model produces very different predictive likelihoods after perturbation should be good candidates for data acquisition to improve the robustness of the current model.

\paragraph{Creating Perturbation for Unlabeled Candidates}

The first step of our PAL acquisition strategy 
finds the distracting sentence from the context of the most similar labeled questions using the embedding of the fine-tuned model.
Specifically, we use the {\em [CLS]}  token embedding from $Encoder_t$ of current fine-tuned model $M_t$ to represent the contexts for all the questions in $\mathcal{D}_{(t)}l$ and $\mathcal{D}_{(t)}u$. We use a K-Nearest-Neighbors (KNN) implementation in order to query the questions with
most similar contexts in $\mathcal{D}_{(t)}l$ for each candidate $u\_i \in \mathcal{D}_{(t)}u$\footnote{The benchmark dataset SQuAD contains many questions with the same context, which we explicitly excluded when choosing the most similar sentence as the distractor.}.
Our distance metric $d(.)$ is Euclidean distance. 
To find the most similar context in $\mathcal{D}_{(t)}l$ for each $u_i$, we select the
top k instead of selecting a predefined threshold. This way, we create a neighborhood set
$\mathcal{N}_{(c)}$ that consists of k most similar contexts in $\mathcal{D}_{(t)}l$ for each $u_i$.
We then break the contexts in $\mathcal{N}_{(c)}$ into sentences, and get the {\em [CLS]} token embeddings of each sentence with $Encoder_t$. 
Similarly, we compute the Euclidean distance between each of these sentences to the context of $u_i$ using embedding, and choose the most similar sentence as the distractor sentence. 
A perturbed instance $u^{'}_i$ is generated by appending the distractor sentence to the original context of $u_i$.

\paragraph{Scoring Robustness to Perturbation} 

In the second step, we  
use the current trained model $\mathcal{M}_{(t)}$ to
obtain the output answer start and end token probabilities for each of the candidate unlabeled question $u_i$ and its corresponding perturbed question $u^{'}_i$.
Then 
we compute the Kullback-Leibler (KL) divergences\footnote{The vanilla Kullback-Leibler divergence is not a symmetric metric, instead, we used the symmetrised divergence: 
$D_{\text{KL}}(P\parallel Q)+D_{\text{KL}}(Q\parallel P)$. \\ \url{https://en.wikipedia.org/wiki/Kullback-Leibler_divergence}
} in the model predictive
probabilities between each ($u_i$, $u^{'}_i$) pair for answer start and end tokens separately, and the negative sum of them is referred as the robustness score $s_{u_i}$ for $u_i$. 
 
\paragraph{Rank Unlabeled Candidates and Select Batch}

We apply these steps to all candidate questions $u_i
\in \mathcal{D}_{(t)}u$ and obtain a score $s_{u_i}$ for each.
A lowest $s_{u_i}$ score indicates that the unlabeled question $u_i$ 
whose predictive probability distributions diverge the most from their predictions after adding the distracting sentence, 
suggesting that it 
is more sensitive to the distracting perturbation, making it a good
candidate for data acquisition to choose to
improve the robustness of the current model. 
To this end, our acquisition function selects $b$ unlabeled questions that have the lowest score $s_{u_i}$
from current unlabeled pool $\mathcal{D}_{(u)}$.

In short, our PAL acquisition strategy finds the distracting sentence from the context of the most similar labelled question using the embedding of the fine-tuned model, and then selects unlabeled questions from $D_u$, whose predictive probability distributions diverge the most from their predictions before adding the distracting sentence measured by Kullback-Leibler divergence.   
    
\section{Evaluation and Results} 
We conduct our evaluation using the Stanford Question Answering Dataset (SQuAD) \citep{rajpurkar2016squad}. 
Before we feed the texts of each question-context pair to the model, we pre-process them with {\em `Tokenizer`} from {\em HuggingFace's \footnote{https://huggingface.co} Transformers} library, which tokenizes the inputs to convert the tokens to their corresponding IDs in the vocabulary.
Then we compute the offset mappings that map tokens to character positions in the context, and save the start and end positions of the answer span as the labels.
We call {\em AutoModelForQuestionAnswering} to initiate a model that
loads the BERT-base encoder with a question-answering header. 
This model predicts the start and the end position of the answer, and calculates the cross-entropy loss by taking the difference between the predicted and labeled start and the end positions.  
At the beginning, all questions in the training dataset are considered to be unlabeled. To start, 1\% of the dataset is selected and labeled for training the model.
Following Algorithm \ref{algorithm:AL-loop},
in each iteration, we continue fine-tuning the model on the newly labelled 10\% of the rest of the unlabeled dataset selected by the active learning acquisition functions. 
This process was repeated until $\mathcal{D}_{(i)}u$ is empty.

\begin{table}[h!]
\centering
\small
\setlength\extrarowheight{3pt}
\begin{tabular}{p{15mm}p{3mm}p{3mm}p{3mm}p{3mm}p{3mm}p{3mm}p{3mm}p{3mm}}
\toprule
\scalebox{.9}[1.0]{AL Strategies}   & 200 & 300 & 400  & 500  & 600 & 700  & 800  & AUC \\ \hline   
Confidence & 52.4 & 66.5 & 71.7 & \textbf{74.8} & \textbf{77.2} & \textbf{77.5} & 79 & 71.3\\
Clustering       & 53.6 & 67.1 & 68.4 & 73.6 & 76.3 & 76.5 & 77.7 & 70.5 \\
Diversity & 50.4 & 65.7  & 71.4 & 71.6 & 75.6 & 74.7 & 78.2 & 70 \\
PAL &      \textbf{57.7} & \textbf{70.1}  & \textbf{72.6} & 74.1 & 76.2 & 78.5 & \textbf{79.9} & \textbf{72.7}\\   
 
\bottomrule
\end{tabular}
\caption{Fine-tuning BERT-base model with various AL acquisition strategies for the question answering task.
The F1 scores are evaluated at every n-th training step (with batch size of 12) on the SQuAD dataset\protect\footnotemark 
}  
\label{tab:al-results}
\end{table}
\footnotetext{SQuAD dataset is a large annotated QA dataset, containing 100,000+ question-answer pairs on 500+ articles. For a context, there are one or more question-answer pairs are associated. For the purpose to demonstrating the benefits of active learning on a smaller annotated dataset in general, we experiment fine-tuning on subset of the whole training dataset, consisting of about 20,000 unique question-answer pairs, one pair per context.}

We experimented with the several common AL strategies and our novel PAL strategy for choosing the instances to label in each iteration.  
Experimental results show that:

\begin{enumerate}
    \item  In general, uncertainty-based strategy outperforms the two other common sampling strategies as it always searches for the ``valuable'' samples around the current decision boundary, but the optimal decision boundary cannot be found unless a certain number of samples have already been labeled, which explains its lower performance at the beginning.
    \item Clustering sampling strategy performs better when the number of labeled samples is very small, while uncertainty-based criterion usually overtakes the clustering strategy afterwards. This is expected and consistent with observations from previous works \citep{zhan2021comparative}.
    The main reason is that the clustering strategy could obtain the entire structure of a database in the beginning stage, but it is insensitive to the data samples that are close to the decision boundary, and requires querying a large number of instances before reaching the optimal decision boundary. 
    \item Both uncertainty-based and clustering sampling strategies can only guarantee their optimal performance over a period of time in the entire AL processes, and the optimal period differs.
    \item The diversity-based sampling is less effective than the other two sampling strategies for the QA task.
    \item Our PAL method is effective and outperforms the commonly used AL acquisition strategies on a fixed annotation budget. 
    
\end{enumerate}

\section{Discussion}

The main challenge in the development of novel AL methods is that there is no objectively superior way to determine how informative each data label is.    
As clustering and diversity-based methods are highly representation-dependent,
there are two main approaches to acquiring data for active learning - uncertainty sampling and representation based sampling. Uncertainty sampling relies on the model's predictive confidence to select difficult examples, while representation based sampling exploits heterogeneity in the feature space.  
Algorithms based on uncertainty may end up choosing uncertain yet uninformative repetitive data, while representation-based methods may tend to select diverse yet easy examples for the model \citep{roy2001toward}.
The two approaches are orthogonal to each other, since uncertainty sampling is usually based on the model's output, while representation exploits information from the input (i.e. feature) space \citep{margatina2021active}.
Our PAL acquisition method utilizes both the input feature (embeddings in the feature space) and model's output (predictive probability distributions) to select the most informative instances. 

Instead of difficulty or heterogeneity, we consider the robustness of the model against perturbation as a signal of informativeness for active learning, by choosing the unlabeled questions where the model produces very different predictive likelihoods after perturbation. 
Hybrid data acquisition functions that combine uncertainty and representation-based sampling have also been presented in several works \citep{yuan2020cold,ru2020active,zhu2008active,shen2004multi}. 
Incorporating our PAL strategy with both uncertainty and representation-based sampling is promising, which
integrates the advantages of multi-criteria as they are different informative dimensions. 
\section{Conclusion}

State-of-the-art textual QA systems are commonly based on a deep machine comprehension model.
Such models are not robust enough and are extremely vulnerable to adversarial examples. To improve system performance, a large amount of labeled training data is often required, 
that is, questions for which the correct answers are known.  
However, a large and high-quality set of labeled training data is not always available, and in practice the acquired labels are not only expensive and also noisy. 
To improve the practical applicability of QA systems, the goal is thus to attain the best gain from this limitation.
Active learning is a machine learning methodology in which a model is trained using a set of most critical labeled data that is selected by the model with certain acquisition strategies, rather than using a set of all available data.  
It is a widely-used training strategy for maximizing predictive performance subject to a fixed annotation budget.  
In this work, we introduce an novel active learning strategy, PAL, for the QA task. The strategy works by sampling unlabeled questions where the model produces very different predictive likelihoods after perturbation with a distractor sentence.
We compare our PAL strategy to three other AL acquisition strategies: model confidence, clustering and question diversity. We used SQuAD as the benchmark dataset for the QA task. 
The experimental results confirm the effectiveness of our PAL strategy, compared to several widely adopted AL acquisition strategies.

\bibliography{acl_latex}
\bibliographystyle{acl_natbib}

\end{document}